\documentclass[10pt,twocolumn,letterpaper]{article}

\usepackage{cvpr}              % To produce the CAMERA-READY version
\usepackage{graphicx}
\usepackage{amsmath}
\usepackage{amssymb}
\usepackage{booktabs}
\usepackage{xcolor}
\usepackage{bm}
\usepackage{float}
\usepackage{diagbox}
\usepackage{tabularx}
\usepackage{multirow}
\usepackage{lipsum}

\usepackage[pagebackref,breaklinks,colorlinks]{hyperref}

\usepackage[capitalize]{cleveref}
\crefname{section}{Sec.}{Secs.}
\Crefname{section}{Section}{Sections}
\Crefname{table}{Table}{Tables}
\crefname{table}{Tab.}{Tabs.}

\def\cvprPaperID{} % *** Enter the CVPR Paper ID here
\def\confName{CVPR}
\def\confYear{2023}

\def\etal{\textit{et~al.~}}		  % and others, and co-workers
\def\eg{\textit{e.g.,~}}               % for example
\def\ie{\textit{i.e.,~}}      % that is, in other words
\def\etc{etc}                 % and other things, and so forth
\newlength\paramargin
\newlength\figmargin

\newlength\secmargin
\newlength\figcapmargin
\newlength\tabcapmargin

\newcommand{\topic}[1]
{
\vspace{1.5mm}\noindent\textbf{#1}
}

\newcommand{\figref}[1]{Figure~\ref{#1}} 
\newcommand{\tabref}[1]{Table~\ref{#1}}
\newcommand{\eqnref}[1]{Equation~\eqref{#1}}

\long\def\ignorethis#1{}

\newbox\jsavebox%

\makeatletter
\newcommand{\providelength}[1]{%
  \@ifundefined{\expandafter\@gobble\string#1}
   {% if #1 is undefined, do \newlength
    \typeout{\string\providelength: making new length \string#1}%
    \newlength{#1}%
   }
   {% else check whether #1 is actually a length
    \sdaau@checkforlength{#1}%
   }%
}

\begin{document}

\title{Benchmarking Robustness in Neural Radiance Fields}
\author{
  Chen Wang$^1$ ~~  Angtian Wang$^2$ ~~ Junbo Li$^3$ ~~ Alan Yuille$^2$ ~~ Cihang Xie$^3$ \vspace{.35em}\\ 
  $^1$ Tsinghua University ~~ $^2$ Johns Hopkins University ~~
    $^3$ UC Santa Cruz \\
}
\maketitle

%%%%%%%%% ABSTRACT
\begin{abstract}
Neural Radiance Field (NeRF) has demonstrated excellent quality in novel view synthesis, thanks to its ability to model 3D object geometries in a concise formulation. However, current approaches to NeRF-based models rely on clean images with accurate camera calibration, which can be difficult to obtain in the real world, where data is often subject to corruption and distortion. In this work, we provide the first comprehensive analysis of the robustness of NeRF-based novel view synthesis algorithms in the presence of different types of corruptions.

We find that NeRF-based models are significantly degraded in the presence of corruption, and are more sensitive to a different set of corruptions than image recognition models. Furthermore, we analyze the robustness of the feature encoder in generalizable methods, which synthesize images using neural features extracted via convolutional neural networks or transformers, and find that it only contributes marginally to robustness. Finally, we reveal that standard data augmentation techniques, which can significantly improve the robustness of recognition models, do not help the robustness of NeRF-based models. We hope that our findings will attract more researchers to study the robustness of NeRF-based approaches and help to improve their performance in the real world.

\end{abstract}

%%%%%%%%% BODY TEXT
\section{Introduction}
Novel View Synthesis (NVS), a long-standing problem in computer vision and computer graphics research, aims to generate photo-realistic images at unseen viewpoints of a 3D scene given a set of posed images. Existing works, including those of image-based rendering, primarily rely on explicit geometry and hand-crafted heuristics~\cite{hedman2016scalable, thies2019deferred, zhou2018stereo}, which require sophisticated design, extensive efforts for data collection and preprocessing and have difficulty in generalizing to new scenes and settings.

\begin{figure}[t]
    \centering
    \includegraphics[width=1.0\linewidth]{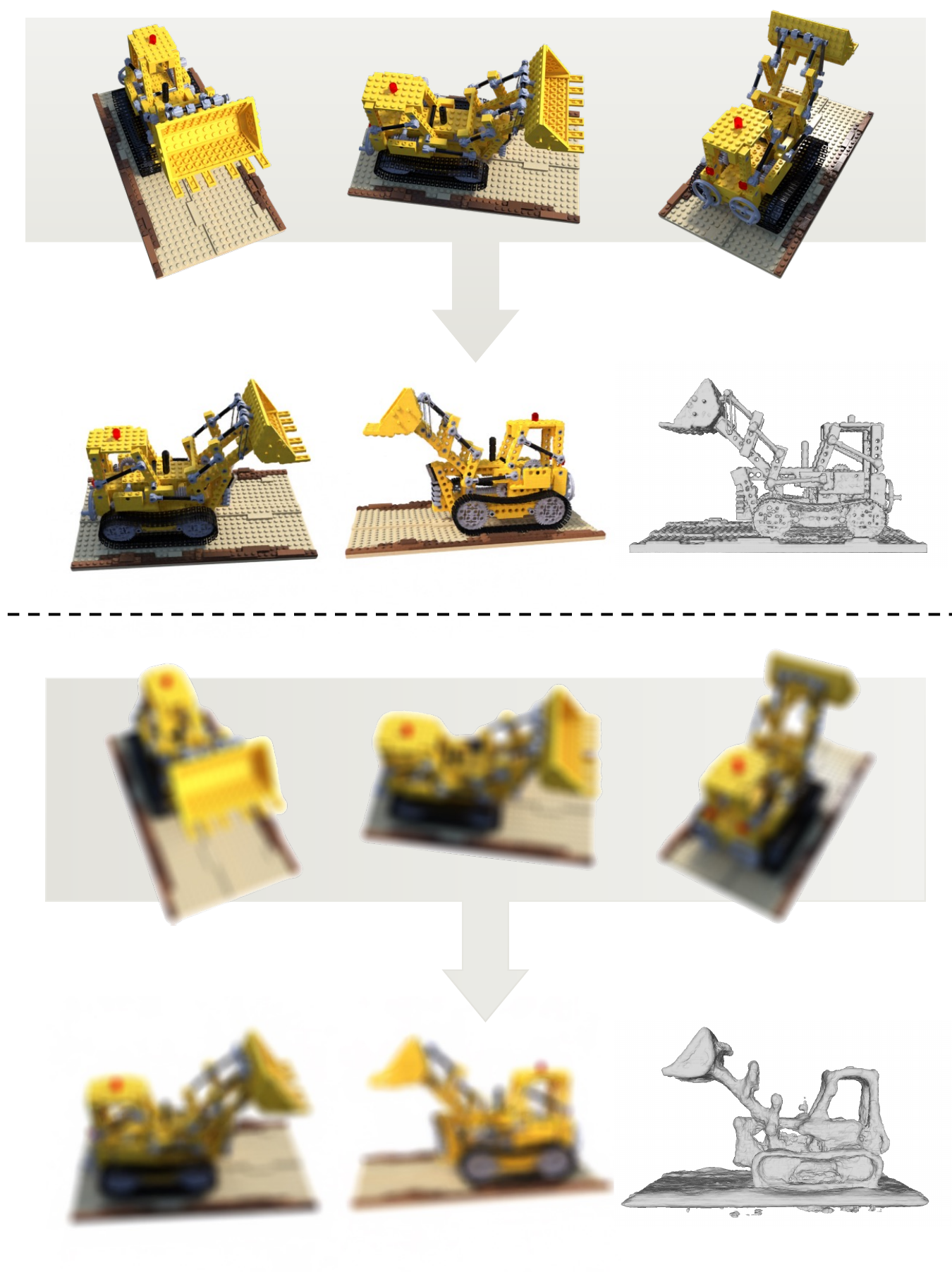}
    \vspace{-2.5em}
    \caption{
    NeRF produces high quality novel view synthesis and accurate surface reconstruction when training on clean images (top). However, corruption on training images will significantly affect both the renderings and geometry of the objects (bottom). Meshes were obtained with marching cubes~\cite{lorensen1987marching}. 
    }
    \vspace{-1em}
    \label{fig:teaser}
\end{figure}

Recently, Neural Radiance Fields (NeRF)~\cite{mildenhall2020nerf} have demonstrated as an effective implicit 3D scene representation over recent years and achieved state-of-the-art performance on NVS by leveraging end-to-end learnable components with 3D geometry context to reconstruct input images. NeRF encodes a scene into a continuous multi-layer perceptron (MLP), which regresses color and density given any 3D location and view direction. Novel views can thus be synthesized through differentiable volumetric rendering from arbitrary viewpoints.

The emergence of NeRF has made NVS more usable in real-world scenarios by reducing the need for tedious preprocessing steps. In practice, we only have to take images with our phones and estimate camera poses with off-the-shelf structure-from-motion (SfM) techniques to train NeRF-based approaches, which can achieve remarkable view synthesis results with accurately calibrated poses and clean images. 
However, to generate photo-realistic images at novel viewpoints, it is necessary to comprehensively model the 3D space, including the scene geometry, illumination, and reflections. This requires capturing local high-frequency details, which can make NeRF-based methods sensitive and fragile to perturbations in the inputs. In addition, NeRF-based methods are often optimized using a pixel-based L2 reconstruction loss, which can lead to overfitting the target scene without prior information. When the inputs are corrupted, such as being compressed in JPEG format or blurred due to motion, the reconstructed scenes may exhibit visible artifacts. Most importantly, corruptions often occur during the real-world capture and preprocessing stages. It may seem evident that they can lead to inaccurate reconstruction, but the more important and interesting question---\emph{how different types and severities of corruption affect the robustness of NeRF}---remains open.

As a step forward, in this paper, we present the first benchmark to comprehensively evaluate the robustness of current NeRF-based methods. Firstly, we construct two benchmarking datasets: \textit{LLFF-C} and \textit{Blender-C}, both of which are the corrupted counterparts of the standard datasets used in NeRF-based methods, and the latter also contains 3D-aware corruptions. Using the metrics we propose, we show that modern NeRF-based models exhibit significant degradation across all types of corruptions, and that there is still a significant opportunity for improvement on both LLFF-C and Blender-C. In addition, difficulties for each corruption type exhibit a totally different nature in NeRF-based methods from recognition tasks. Especially, for generalizable methods that use image features to assist neural rendering, we find that scaling the feature encoder does not enhance robustness. We also show that fine-tuning a pretrained generalizable model leads the model to overfit the ``incorrect'' data, sometimes resulting in worse synthesis quality than using the pretrained model directly.

In summary, the contributions of our paper include: 
\begin{itemize}
    \item We present the first framework for benchmarking and assessing the robustness of NeRF-based systems to visual corruptions. 

    \item Our findings demonstrate that current NeRF-based methods perform poorly under corruptions, including those \textit{generalizable} ones.

    \item We systematically analyze and discuss the robustness of NeRF-based methods under various corruptions, feature encoder design, \etc. 

    \item We find that standard image data-augmentation techniques do not improve the robustness of novel view synthesis that relies on cross-view consistency.
\end{itemize}

\section{Related Work}
\topic{Novel view synthesis with NeRF.} Traditional image-based rendering methods generate novel views by warping and blending input frames~\cite{gortler1996lumigraph, levoy1996light}, and learning-based methods predict blending weights through neural networks or hand-crafted heuristics~\cite{hedman2018deep, riegler2021stable, riegler2020free}. Different from these, geometry-based methods render images through an explicit 3D model, \eg Thies \etal~\cite{thies2019deferred} stored neural textures on 3D meshes and rendering with standard graphics pipeline. Other 3D proxies such as point clouds~\cite{ruckert2022adop, aliev2020neural}, voxel grids~\cite{penner2017soft, kalantari2016learning}, multi-plane images~\cite{mildenhall2019local, srinivasan2019pushing, zhou2018stereo} are also used. However, these approaches often require large amounts of data and memory to produce satisfying results.

Recently, neural fields~\cite{xie2022neural} leverages an MLP to represent 3D shapes or scenes by encoding them into signed-distance, occupancy, or density fields. Among them, NeRF~\cite{mildenhall2020nerf} demonstrated remarkable results for novel view synthesis with posed images. NeRF uses a coordinate-based MLP representation and obtains color by differentiable volumetric rendering. The optimization of NeRF can be simply done by minimizing the photometric loss at training camera viewpoints. Although NeRF has been investigated in several aspects, \eg generation~\cite{niemeyer2021giraffe}, 3D recontruction~\cite{wei2021nerfingmvs, wang2021neus}, 3D super-resolution~\cite{wang2021nerf}, in this work, we still focus on the novel view synthesis task.

\topic{Robustness benchmarks.} The robustness of deep learning models is crucial for real-world applications, and its assessment has received growing attention in recent years~\cite{hendrycks2019benchmarking, recht2019imagenet, kar20223d}. ImageNet-C~\cite{hendrycks2019benchmarking} benchmark, as one of the pioneering works, evaluates image classifiers' robustness under simulated image corruptions such as motion blur and jpeg compression. Imagenet-V2~\cite{recht2019imagenet} creates new test sets for ImageNet and CIFAR-10 and evaluates the accuracy gap caused by natural distribution shift. Specifically for object recognition, ObjectNet~\cite{barbu2019objectnet} presents a real-world test set containing objects with random backgrounds, rotations, and imaging viewpoints.  ImageNet-A~\cite{hendrycks2021natural} and ImageNet-R~\cite{hendrycks2021many} further propose additional benchmarks for natural adversarial examples and abstract visual renditions like image style, camera operation, and geographic location \etc.

Researchers have also examined benchmarks beyond image classification. RobustNav~\cite{chattopadhyay2021robustnav} quantifies the performance of embodied navigation agents when exposed to both visual and dynamic corruptions. Ren \etal~\cite{ren2022benchmarking} provide a taxonomy of common 3D corruptions and identify the atomic corruptions for point clouds for the first time, followed by an evaluation of existing point cloud classification models. More recently, Kar~\cite{kar20223d} proposes 3D common corruptions that resemble real-world ones by integrating geometry information \ie depth, into the corruption process. However, a comprehensive benchmark for evaluating the robustness of NeRF is still lacking.

\topic{Improving model robustness.} Adversarial training~\cite{madry2017towards} is a common method used to protect models from corruption. For example, Xie \etal~\cite{xie2020adversarial} show that adversarial perturbations can be used as data augmentation and improve image classification accuracy. Similar conclusions are also reached in natural language processing~\cite{wang2019improving}.

On the other hand, data augmentation can significantly improve generalization performance, and many works augment input images to boost recognition. Mixup~\cite{zhang2017mixup} enforces neural networks to favor linear behavior between training examples by creating convex interpolated samples. Similarly, CutMix~\cite{yun2019cutmix} also mixes two images, but embeds one in a part of another. Augmix~\cite{hendrycks2019augmix} creates compositions of multiple augmented samples that preserve original semantics and statistics. However, mixup~\cite{zhang2017mixup} and cutmix~\cite{yun2019cutmix} are not suitable for NeRF augmentation as they disturb the cross-view constraint required by NeRF, while the efficacy of augmix~\cite{hendrycks2019augmix} remains to be studied.

To increase the robustness of NeRF, Aug-NeRF~\cite{chen2022aug} firstly proposes a triple-level augmentation training pipeline that is robust to noisy inputs. Some works tackle specific corruptions, \eg Deblur-NeRF~\cite{ma2022deblur} for blurred images, NAN-NeRF~\cite{pearl2022nan} for burst noise. Our work differs from them in that we aim to evaluate the robustness of standard NeRF methods and seek ways to improve them.

\begin{figure*}
    \includegraphics[width=1.0\linewidth]{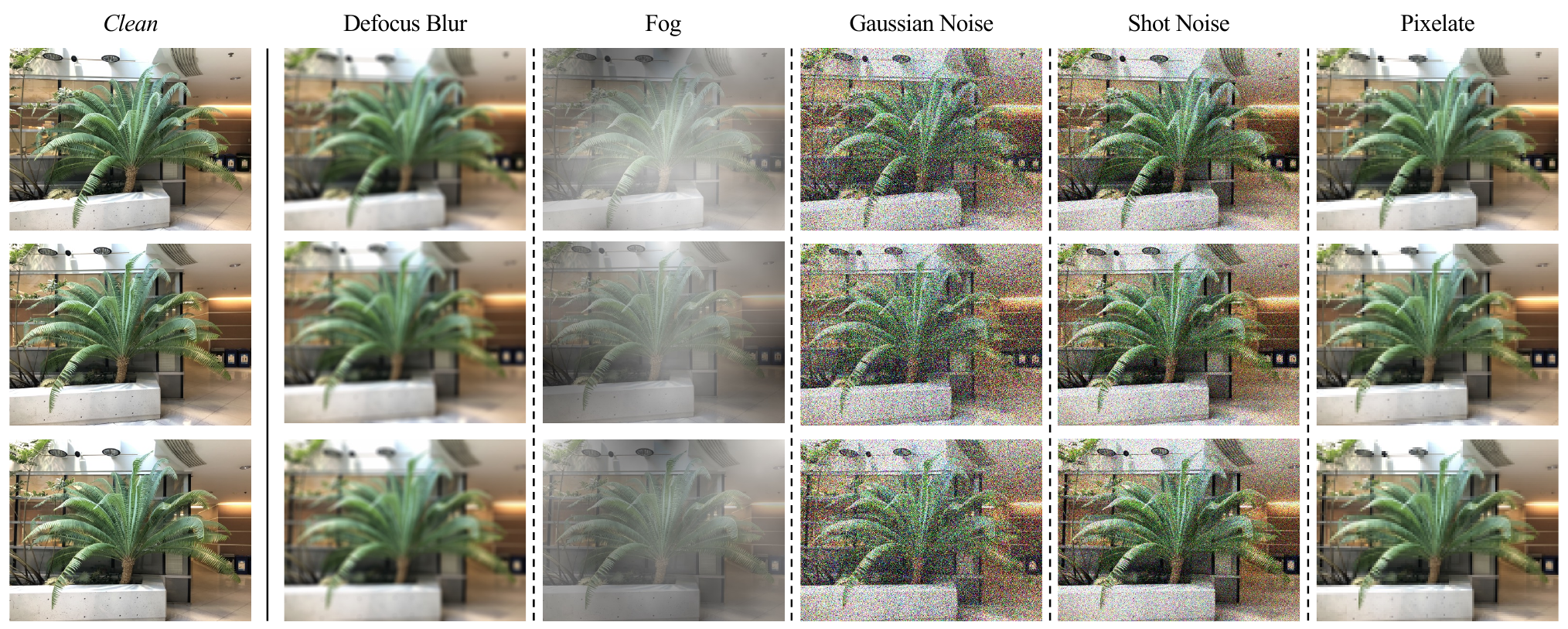}
    \caption{Examples of our proposed LLFF-C on the \textit{fern} scene under five corruption types at the severity of 2.}
    \label{fig:llffc}
\end{figure*}

\begin{figure}
    \includegraphics[width=1.0\linewidth]{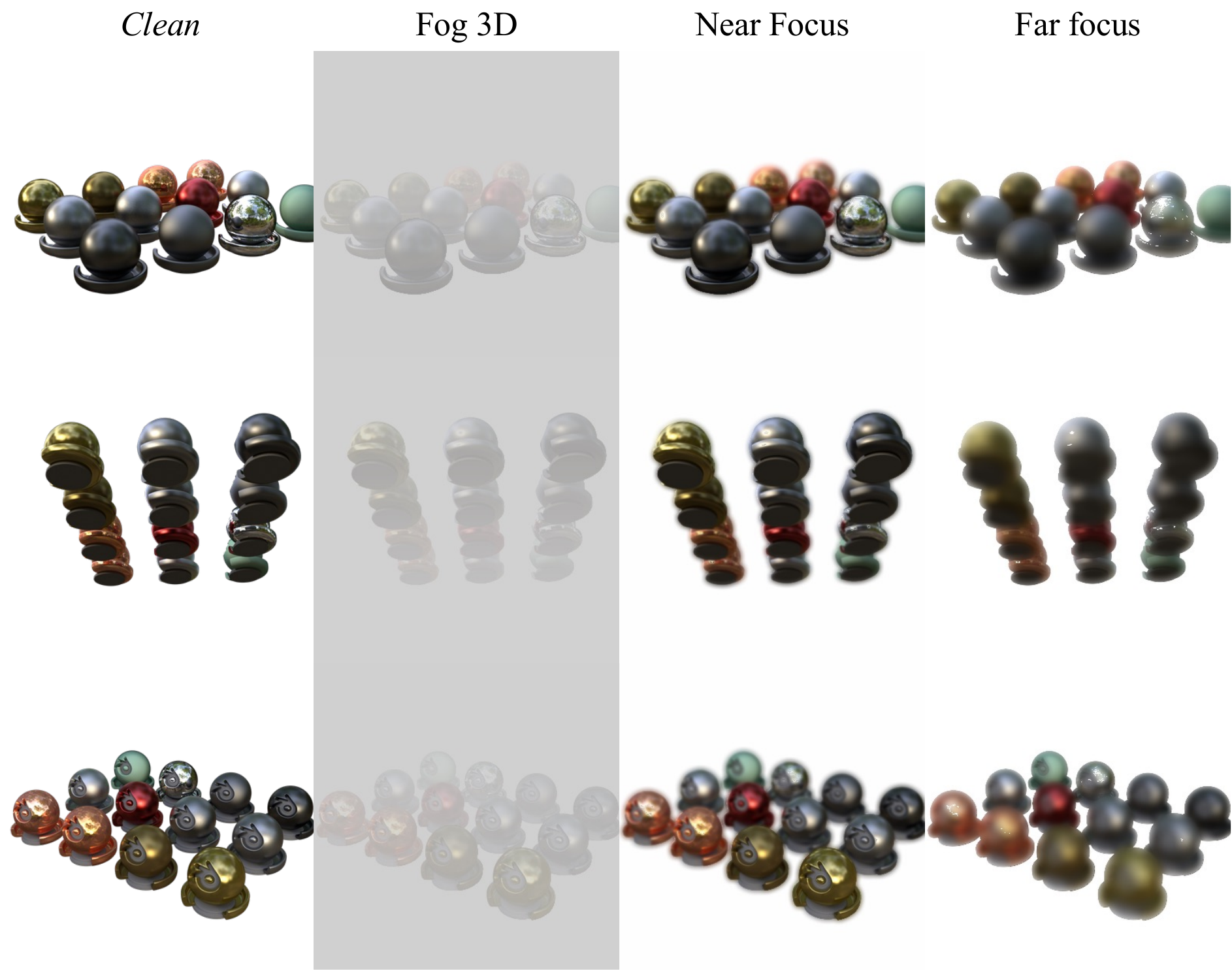}
    \caption{Examples of our proposed Blender-C (left) on the \textit{materials} scene under 3D-aware corruptions, from which we can see that the same corruption has varying effects across scene depth.}
    \label{fig:blenderc}
\end{figure}

\section{Background}
We present a benchmark for evaluating the robustness of rendering models that utilize NeRF~\cite{mildenhall2020nerf}. This section provides an overview of NeRF and our classification of NeRF-based methods.
\label{sec:bg}
\subsection{NeRF for Novel View Synthesis}
\label{subsec:nerf-based}
NeRF represents a 3D scene as a continuous function, which takes as inputs a 5D vector containing 3D position $\mathbf{x} = (x, y, z)$ and viewing direction $\mathbf{d} = (\boldsymbol{\theta}, \boldsymbol{\phi})$, and outputs the corresponding radiance $\mathbf{c}(\mathbf{x}, \mathbf{d}) = (r, g, b)$ with volume density $\sigma(\mathbf{x})$. NeRF is typically parameterized as an MLP $f: (\mathbf{x}, \mathbf{d}) \rightarrow (\mathbf{c}, \sigma)$.

NeRF is an emission-only model, \ie the color of a pixel only depends on the radiance along the viewing ray. Therefore, according to volume rendering \cite{kajiya1984ray}, the color along the camera ray $\mathbf{r}(t) = \mathbf{o} + t\mathbf{d}$ that shots from the camera center $\mathbf{o}$ in direction $\mathbf{d}$ can be calculated via standard volume rendering, the discrete format is expressed as the following:
\begin{equation}F
    \hat{\mathbf{C}}(\mathbf{r}) = \sum_{i=1}^{N}T_{i}(1 - \mathrm{exp}(-\sigma_{i}\delta_{i}))c_{i}, 
\label{equ:render}
\end{equation}
\begin{equation}
    T_{i} = \mathrm{exp}(-\sum_{j=1}^{i-1}\sigma_{j}\delta_{j}),
\end{equation}
where $N$ is the number of sampled points along the ray, $\delta_{i} = t_{i+1}-t_{i}$ is the distance between two adjacent samples, $c_{i}$ and $\sigma_{i}$ are the per-point radiance and density, and $T_{i}$ denotes the accumulated transmittance.

NeRF is trained to minimize the mean-squared error (MSE) between the predicted renderings and the corresponding ground-truth color:
\begin{equation}
    \mathcal{L}_{\mathrm{MSE}} = \frac{1}{|\mathcal{R}|}\sum_{r \in \mathcal{R}}\| \hat{\mathbf{C}}(\mathbf{r}) - \mathbf{C}(\mathbf{r}) \|^{2},
    \label{equ:mse}
\end{equation}
where $\mathcal{R}$ denotes the batch of rays randomly sampled from all training images or one specific image. $\hat{\mathbf{C}}(r)$ and $\mathbf{C}(r)$ are the ground truth and output color of ray $r$. This per-pixel optimization lacks holistic spatial understanding and might make NeRF sensitive to disturbance in pixel values.

After the emergence of NeRF, several other methods based on NeRF have been proposed for novel view synthesis. Although they differ in many aspects, \eg sampling strategy, positional encoding, network architecture \etc, all of them aggregate colors and densities of discontinuous points along viewing rays via differentiable volumetric rendering to synthesize novel views, which are named \textit{NeRF-based} methods. Non-NeRF-based neural rendering methods obtain pixel colors using explicit representations \cite{ruckert2022adop, wang2022voge}, \eg point clouds or surface-based implicit representation \cite{yariv2020multiview} without volume densities.

\subsection{Two genres of NeRF-based methods}
\label{subsec:genres}
We further divide NeRF-based methods into two categories: \textit{scene-specific} and \textit{generalizable}. Scene-specific methods such as~\cite{mildenhall2020nerf, barron2021mip} optimize a single model from scratch with a set of training images of one scene, leading to different network parameters for each scene. In contrast, generalizable methods~\cite{wang2021ibrnet, chen2021mvsnerf, yu2021pixelnerf} firstly train a model on a dataset containing hundreds of 3D scenes, then directly infer or fine-tune a few steps on a single testing scene. Specifically, generalizable methods contain CNN or transformer encoders $\bm{F}$ to extract image features from inputs $\mathcal{I}_{i}, i = 1, 2, ... n$:
\begin{equation}
\bm{f}_{i} = \bm{F}({\mathcal{I}_{i}}),
\end{equation}
For a 3D point $\bm{p}$, image features $\bm{f}_{i}$ across input views are aggregated by the function $\mathcal{A}$. $\mathcal{A}$ is quite different in different methods, but often it consists of a series of operations: perspective projection $\bm{p}$ into input each image, image feature interpolation, and multi-view feature fusion (\eg cost volume, pooling or simple concatenation):
\begin{equation}
\bm{f}_{\bm{p}} = \mathcal{A}(\bm{f}_{i}; \bm{p}),
\end{equation}
The features are further decoded to the color $\mathbf{c}_{\bm{p}}$ and density $\sigma_{\bm{p}}$ of $\bm{p}$:
\begin{align}
\mathbf{c}_{\bm{p}} &= \mathcal{D}_{c}(\bm{f}_{\bm{p}}; \bm{z}_{c}), \\
\sigma_{\bm{p}} &= \mathcal{D}_{\sigma}(\bm{f}_{\bm{p}}; \bm{z}_{\sigma}),
\end{align}
where $\bm{z}_{c}$ and $\bm{z}_{\sigma}$ are optional auxiliary vectors (\eg visibility) to enhance decoding, $\mathcal{D}_{c}$ and $\mathcal{D}_{\sigma}$ denote the decoding networks (mostly MLPs) for color and density respectively ($\mathcal{D}_{c}$ and $\mathcal{D}_{\sigma}$ sometimes share the parameters). With colors and densities for queried 3D points, the final pixel color can be similarly obtained using \eqnref{equ:render}.

\section{Corruptions and Test Suite}
\label{sec:tax}
In this section, we present a detailed description of the structure of our benchmark. Our study focuses on the impact of corruption on NeRF-based models, and we primarily conduct experiments on NeRF-based novel view synthesis methods. However, our test suite can also be directly applied to recent non-NeRF-based methods, such as the one presented in \cite{suhail2022generalizable}, as described in Section \ref{sec:exp}. Additionally, the test suite can be easily adapted to other tasks that involve NeRF, \eg relighting, reconstruction, and video synthesis.

The goal of our work is to act as a foundation for building robust NeRF-based systems in the real world. Unlike previous benchmarks~\cite{hendrycks2019benchmarking}, each data chunk in ours is not a single image, but a 3D scene comprising multi-view images of one scene. \textit{Scene-specific} methods are optimized directly on ``corrupted" unseen scenes, while \textit{generalizable} methods have already been pretrained on clean training scenes and then tested or finetuned on corrupted target scenes. To avoid confusion, in this paper, \textit{training set} only refers to the 3D scenes used for pretraining generalizable methods, while \textit{target training set} and \textit{target testing set} refer to the training and testing images for a specific target scene or object. The corruptions are drawn from a predefined set which we will elaborate in the following.

\subsection{Corruptions}
Similar to images, scene corruptions are artifacts that degrade the quality of a system's RGB observations. The corruptions we include have the following characteristics: (1) the target training set and target testing set have the same lighting conditions; and (2) corrupted images and their clean counterparts are taken from the same camera poses, with only the content altered. We provide around ten types (the exact number depends on the dataset) of corruptions \eg gaussian noise and fog, mainly from those proposed in \cite{hendrycks2019benchmarking} and exclude those that do not meet the above requirements. Each corruption has three levels of severity ($1\xrightarrow{}3$) indicating an increase in the degree of degradation.

\subsection{Datasets}
\topic{LLFF-C.} LLFF~\cite{mildenhall2019local} consists of 8 real-world scenes that contain mainly forward-facing images. Each of the eight images is held out as the target testing set. We corrupt the target training set in each scene with 9 corruptions to create the LLFF-C test suite (see \figref{fig:llffc} for an example).

\topic{Blender-C.} Blender-C is constructed based on the Blender dataset (also known as NeRF-Synthetic)~\cite{mildenhall2022nerf} that contains 8 detailed synthetic objects with 100 images taken from virtual cameras arranged on a hemisphere pointed inward. Inspired by \cite{kar20223d}, we specifically create 3D-aware fog, near focus, and far focus on replacing their 2D counterparts for Blender-C (see \figref{fig:blenderc}). For example, in 3D-aware fog, pixels far from the camera are occluded to a higher extent. The RGB images and the corresponding depth maps are rendered with the official blend files for 3D-aware corruptions. We use 100 images as the target training set and 25 images as the target testing set for each scene.

\subsection{Task and Metrics}
The task of our benchmark is novel view synthesis learned from images of a 3D scene. The standard procedure for evaluating the performance of novel view synthesis methods is to compare the ground truth images and predicted images at testing viewpoints with the three mostly used metrics: Peak Signal-to-Noise Ratio (PSNR) and Structural Similarity Index Measure (SSIM)~\cite{wang2003multiscale} and LPIPS~\cite{zhang2018unreasonable}. For scene-specific models, we directly provide a corrupted target training set for each model. Generalizable methods are trained on clean training scenes but perform inference or finetuning on corrupted target scenes.

Inspired by the 2D common corruptions on images~\cite{hendrycks2019benchmarking} and point clouds~\cite{ren2022benchmarking}, the first step for evaluation is to optimize (inference or finetune for generalizable methods) a model $f$ with a set of clean images and compute the relevant metrics $m_{\mathrm{clean}}^{f}, m \in M$ at the target testing set. Next, the same process will be repeated, but with corrupted target training set for each corruption type $c$ and severity $s$. We then calculated the metrics again,  which are denoted by $m_{\mathrm{c, s}}^{f}, m \in M$. Thus, the corruptions metric (CM) is defined as the mean metric over severities:
\begin{equation}
\mathrm{CM}_{c,m} = \frac{1}{3}\sum_{s=1}^{3}m_{c, s},
\label{equ:cm}
\end{equation}

Different from classification benchmarks, we do not use another \textit{baseline} model as the denominator in \eqnref{equ:cm} because it washes out the models' absolute performance on the corrupted data. Mean CM is thus the average of CM over all the corruption types:
\begin{equation}
\mathrm{mCM}_{m} = \frac{1}{N}\sum_{c}\mathrm{CM}_{c,m},
\end{equation}
where $N$ is the number of corruption types.

While $\mathrm{CM}$ and $\mathrm{mCM}$ measure the absolute robustness of NeRF-based models, we are also interested in their relative performance drop, \ie the amount a model degrades from clean inputs to corrupted ones, defined by Relative mCM as the following:
\begin{equation}
\mathrm{RCM}_{c,m} = \frac{1}{3}\sum_{s=1}^{3}\frac{|m_{\mathrm{clean}} - m_{c, s}|}{m_{\mathrm{clean}}},
\label{equ:rcm}
\end{equation}

\begin{equation}
\mathrm{RmCM}_{m} = \frac{1}{N}\sum_{c}\mathrm{RCM}_{c,m},
\label{equ:rmcm}
\end{equation}
where $m \in M$. We use absolute $|m_{\mathrm{clean}} - m_{c, s}|$ in $\mathrm{RCM}_{c,m}$ considering that PSNR and SSIM are higher the better, while LPIPS is lower the better.

\section{Experiment}
\subsection{Setup}
\label{sec:exp}
We benchmark a total of seven methods, all of which are representative of recent novel view synthesis works. For scene-specific methods, we include NeRF~\cite{mildenhall2020nerf}, MipNeRF~\cite{barron2021mip}, and the network-free method Plenoxels~\cite{fridovich2022plenoxels}. For generalizable methods, we experiment with IBRNet~\cite{wang2021ibrnet} and MVSNeRF~\cite{chen2021mvsnerf}. We also include Generalizable Patch-Based Neural Rendering (GPNR)~\cite{suhail2022generalizable}, the latest pure transformer-based architecture, as we were interested in its robustness compared to other methods. We omit PixelNeRF~\cite{yu2021pixelnerf} due to its poor performance on datasets other than its testing set, DTU.

For all of the methods, we use publicly available codebases and checkpoints, re-training some as necessary. For IBRNet~\cite{wang2021ibrnet} and MVSNeRF~\cite{chen2021mvsnerf}, we present results for both direct inference and fine-tuning. The details of dataset processing, training, etc., for each method are included in the supplementary material.

\begin{table*}[htbp]
\resizebox{\linewidth}{!}{
\begin{tabular}{l|l|ccc|ccc|c|cc}
\toprule[0.5pt]
 & & \multicolumn{3}{c|}{Noise} & \multicolumn{3}{c|}{Blur} & \multicolumn{1}{c|}{Weather} & \multicolumn{2}{c}{Digital} \\
\multicolumn{1}{c|}{}  & \multicolumn{1}{c|}{Clean} & Gauss. & Shot & Impulse & Defocus & Glass & Motion & Fog & Pixel & JPEG \\ \midrule[0.5pt]
NeRF & 27.68/0.151 & \multicolumn{1}{l}{24.32/0.297} & \multicolumn{1}{l}{23.74/0.300} & \multicolumn{1}{l|}{24.04/0.297} & \multicolumn{1}{l}{20.84/0.501} & \multicolumn{1}{l}{22.26/0.372} & \multicolumn{1}{l|}{19.20/0.483} & \multicolumn{1}{l|}{11.87/0.528} & \multicolumn{1}{l}{24.28/0.302} & \multicolumn{1}{l}{25.29/0.288} \\ 
MipNeRF & \multicolumn{1}{c|}{27.69/0.159} & 23.96/0.340 & 23.34/0.345 & 23.69/0.341 & 20.86/0.510 & 22.24/0.380 & 19.10/0.507 & 12.03/0.545 & 24.22/0.316 & 22.13/0.257 \\ 
\multicolumn{1}{l|}{Plenoxel} & \multicolumn{1}{c|}{27.45/0.100} & 20.72/0.480 & 20.31/0.486 & 17.71/0.494 & 20.45/0.509 & 21.91/0.373 & 19.20/0.463 & 12.44/0.660 & 23.81/0.289 & 24.78/0.300 \\ 
\multicolumn{1}{l|}{MVSNeRF} & \multicolumn{1}{c|}{17.03/0.409} & 16.78/0.545 & 16.50/0.551 & 16.72/0.550 & 17.13/0.576 & 17.33/0.500 & 16.30/0.558 & 12.97/0.567 & 17.65/0.448 & 17.37/0.465 \\ 
\multicolumn{1}{l|}{MVSNeRF (ft)} & \multicolumn{1}{c|}{23.94/0.244} & 21.51/0.442 & 21.17/0.443 & 21.41/0.441 & 20.58/0.517 & 21.62/0.408 & 19.42/0.490 & 13.12/0.606 & 22.70/0.353 & 23.15/0.349 \\ 
\multicolumn{1}{l|}{IBRNet} & \multicolumn{1}{c|}{25.71/0.158} & 21.88/0.452 & 21.36/0.455 & 22.05/0.434 & 20.54/0.514 & 21.77/0.392 & 19.39/0.453 & 13.74/0.453 & 23.35/0.311 & 23.91/0.338 \\ 
\multicolumn{1}{l|}{IBRNet (ft)} & \multicolumn{1}{c|}{27.80/0.112} & 24.03/0.321 & 21.33/0.451 & 23.83/0.322 & 19.67/0.517 & 21.84/0.395 & 19.45/0.456 & 13.33/0.501 & 22.82/0.341 & 24.67/0.311 \\ 
\multicolumn{1}{l|}{GPNR} & \multicolumn{1}{c|}{24.58/0.210} & 21.56/0.481 & 21.11/0.482 & 21.43/0.476 & 20.31/0.530 & 21.38/0.418 & 19.13/0.483 & 12.83/0.569 & 22.82/0.341 & 23.34/0.346  \\ \bottomrule[0.5pt]
\end{tabular}
}
\caption{PSNR$\uparrow$ / LPIPS$\downarrow$ results for clean and corrupted data on LLFF-C, ft indicates results after fine-tuning for generalizable methods.}
\label{tab:llffc}
\end{table*}

\begin{table*}[htbp]
\resizebox{\linewidth}{!}{
\begin{tabular}{l|l|ccc|cccc|c|cc}
\toprule[0.5pt]
 & & \multicolumn{3}{c|}{Noise} & \multicolumn{4}{c|}{Blur} & \multicolumn{1}{c|}{Weather} & \multicolumn{2}{c}{Digital} \\
\multicolumn{1}{c|}{}  & \multicolumn{1}{c|}{Clean} & Gauss. & Shot & Impulse & Near Focus & Far Focus & Glass & Motion & Fog & Pixel & JPEG \\ \midrule[0.5pt]
NeRF & 30.98/0.071 & \multicolumn{1}{l}{22.02/0.163} & \multicolumn{1}{l}{19.21/0.208} & \multicolumn{1}{l|}{23.75/0.167} & \multicolumn{1}{l}{26.99/0.127} & \multicolumn{1}{l}{24.81/0.158} & \multicolumn{1}{l}{22.48/0.210} & \multicolumn{1}{l|}{20.73/0.233} & \multicolumn{1}{l|}{15.44/0.200} & \multicolumn{1}{l}{26.81/0.138} & \multicolumn{1}{l}{28.34/0.117} \\ 
MipNeRF & \multicolumn{1}{c|}{33.34/0.061} & 22.68/0.149 & 
 19.59/0.211 & 24.94/0.178 & 27.92/0.099 & 25.93/0.121 & 22.49/0.208 & 21.04/0.211 & 15.52/0.184 & 27.66/0.118 & 29.74/0.098 \\ 
\multicolumn{1}{l|}{Plenoxel} & \multicolumn{1}{c|}{32.94/0.035} & 20.53/0.556 & 17.68/0.603 & 22.41/0.461 & 27.68/0.100 & 25.73/0.116 & 22.28/0.210 & 21.10/0.225 & 15.62/0.477 & 26.56/0.130 & 28.58/0.115  \\ 
\multicolumn{1}{l|}{MVSNeRF} & \multicolumn{1}{c|}{19.56/0.288} & 15.68/0.559 & 14.26/0.603 & 15.36/0.599 & 19.17/0.310 & 18.98/0.322 & 18.24/0.357 & 17.21/0.364 & 14.11/0.467 & 19.28/0.316 & 19.26/0.314 \\ 
\multicolumn{1}{l|}{MVSNeRF (ft)} & \multicolumn{1}{c|}{23.28/0.199} & 17.72/0.512 & 16.12/0.561 & 18.25/0.538 & 18.25/0.538 & 22.87/0.217 & 22.58/0.208 & 21.38/0.255 & 14.33/0.406 & 22.87/0.220 & 23.03/0.217 \\ 
\multicolumn{1}{l|}{IBRNet} & \multicolumn{1}{c|}{27.15/0.143} & 20.45/0.489 & 18.16/0.535 & 21.49/0.499 & 24.98/0.142 & 23.80/0.212 & 21.74/0.270 & 20.64/0.254 & 15.28/0.237 & 24.87/0.120 & 25.64/0.120 \\ 
\multicolumn{1}{l|}{IBRNet (ft)} & \multicolumn{1}{c|}{29.91/0.081} & 20.97/0.483 & 19.97/0.496 & 22.36/0.489 & 24.88/0.126 & 22.72/0.197 & 21.42/0.203 & 19.95/0.210 & 15.17/0.256 & 23.94/0.177 & 23.81/0.186 \\ 
\bottomrule[0.5pt]
\end{tabular}
}
\caption{PSNR$\uparrow$ / LPIPS$\downarrow$ results for clean and corrupted data on Blender-C, ft indicates results after fine-tuning for generalizable methods.}
\label{tab:blenderc}
\end{table*}

\subsection{Results and Findings}
\begin{figure*}[htbp]
    \centering
    \includegraphics[width=1.0\linewidth]{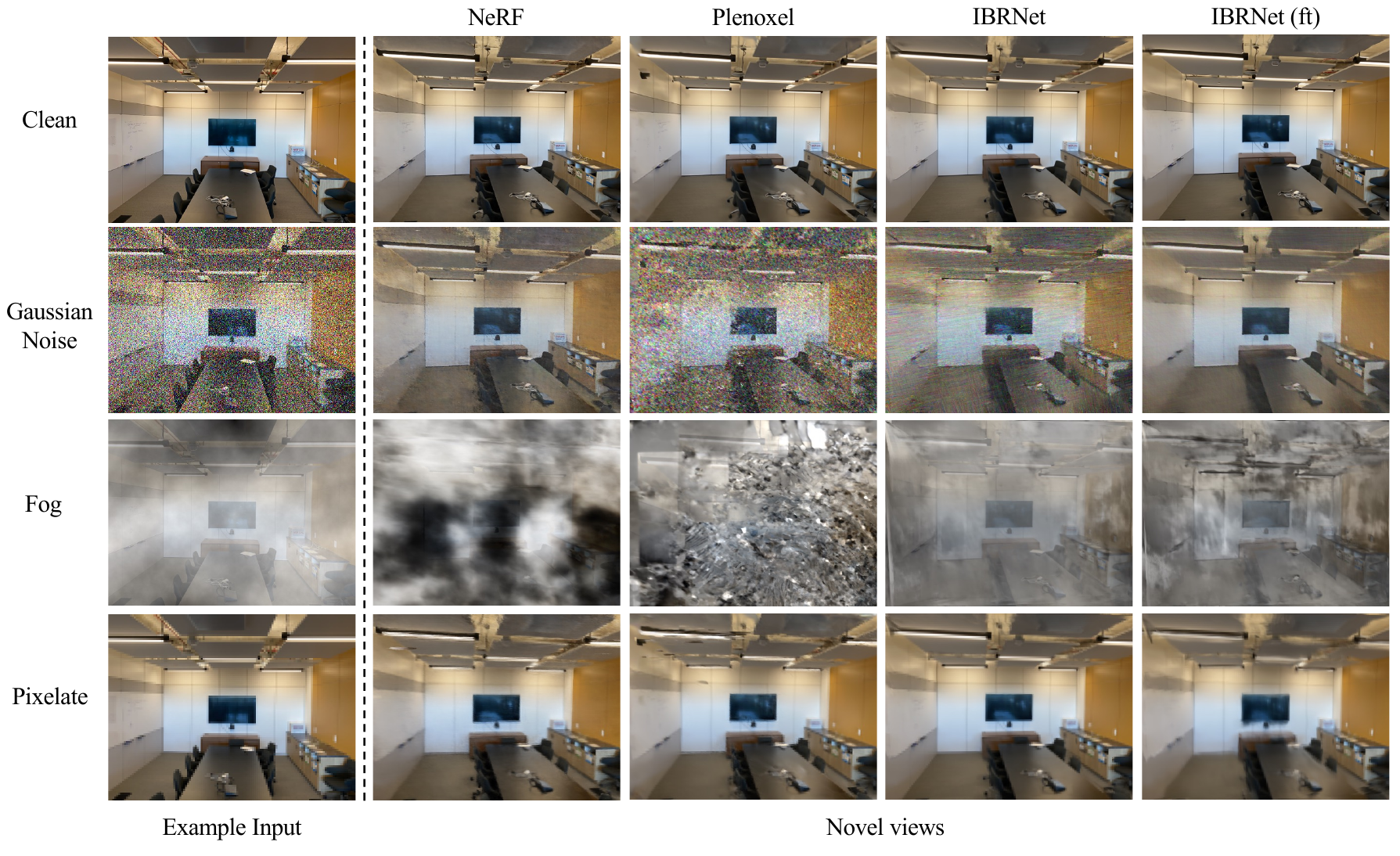}
    \vspace{-2em}
    \caption{Qualitative results across corruptions (severity = 3) and methods on the \textit{room} scene. Note we hereby use multiple input images for the scene and only show one example. We can see that each method behaves differently under these corruptions. We encourage the readers to zoom in for a better inspection.}
    \label{fig:rendering}
\end{figure*}

We report the $\mathrm{CM}$ values of PSNR and LPIPS on LLFF-C and Blender-C at \tabref{tab:llffc} and \tabref{tab:blenderc}. \figref{fig:rendering} shows part of the qualitative results. Other results, including $\mathrm{CM}$ values of SSIM and detailed $\mathrm{RmCM}$ can be found in the supplementary materials. 

\begin{figure}[htbp]
    \centering
    \includegraphics[width=1.0\linewidth]{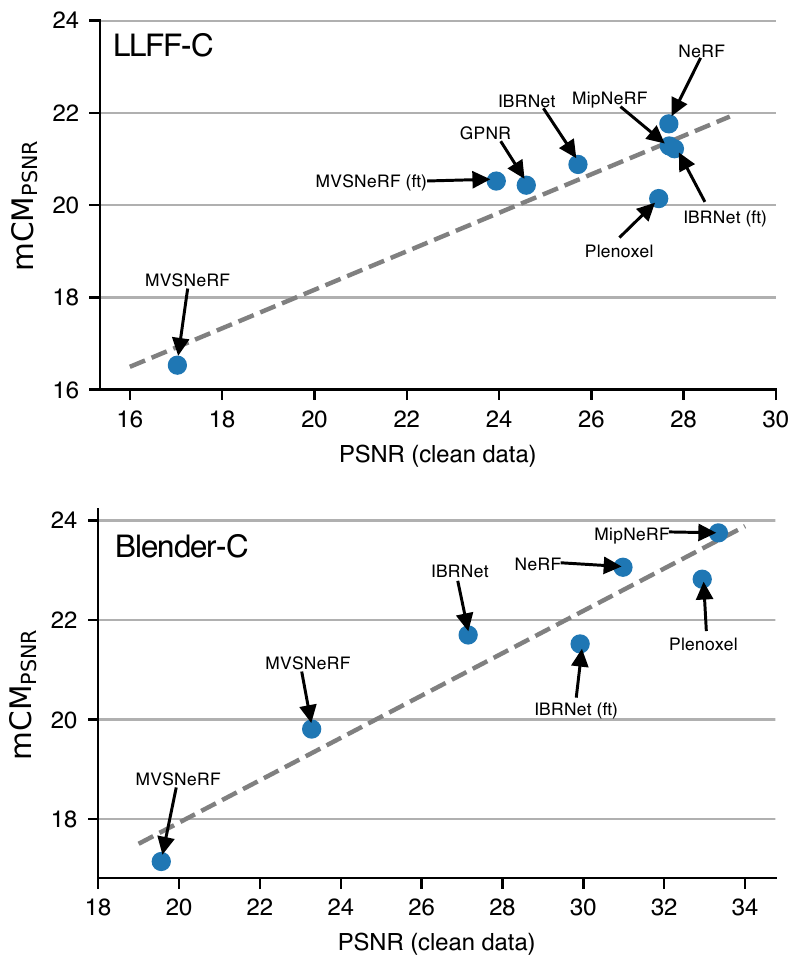}
    \caption{Robustness (mCM) of PSNR values on LLFF-C and Blender-C.}
    \label{fig:trendllffc}
\end{figure}

Generally, all state-of-the-art methods suffer a performance drop from the clean setting across all corruptions. As seen in \figref{fig:trendllffc}, methods that achieve better quality on clean data mostly excel in $\mathrm{mCM}$, with scene-specific models more robust than generalizable ones. However, it seems that the corruption robustness is mainly explained by the model's original ability for scene representation since, by looking at $\mathrm{RmCE}$, no models have demonstrated a remarkable ability to resist corruption.

In terms of relative robustness, Plenoxel~\cite{fridovich2022plenoxels} and IBRNet (ft)~\cite{wang2021ibrnet} have the highest $\mathrm{RmCM}$ on LLFF-C ($25.5\%/331\%$ and $25.7\%/271\%$ for PSNR/LPIPS) and Blender-C ($31.7\%/751\%$ and $28.1\%/248\%$ for PSNR/LPIPS). Moreover, as a voxel-based approach, Plenoxel~\cite{fridovich2022plenoxels} sometimes consumes much higher GPU memory on corrupted data than usual due to incorrect density distributions in the empty space of optimized 3D scenes. Also, from \figref{fig:rendering}, we can find it struggles with Gaussian Noise and Fog. This reveals that explicit representation might not be suitable for highly corrupted situations. MVSNeRF~\cite{chen2021mvsnerf} has the lowest $\mathrm{RmCM}$ on both datasets because its $m_{\mathrm{clean}}$ is fairly low and leaves little room for degradation. Generalizable methods without finetuning are more relatively robust, maybe due to their prior knowledge learned from massive training data.

Not all corruptions are equally severe. For example, \textit{Pixelate} and \textit{JPEG Compression} only lead to an absolute drop of PSNR in less than 15\%. However, \textit{Fog} is the hardest of all methods, resulting in a nearly 50\% absolute drop in PSNR for all methods except MVSNeRF~\cite{chen2021mvsnerf}. The main reason for this huge discrepancy is that the \textit{Pixelate} and \textit{JPEG} have limited influence on original inputs, while \textit{Fog} corruption leads to a drastic change in images, \ie occludes the objects with fog (see \figref{fig:llffc}). The difficulties of recognition tasks and 3D reconstruction can also vary. From \cite{hendrycks2019benchmarking}, for image classification, \textit{Fog} is the easiest corruption type, while \textit{Glass Blur} is the hardest.

With regard to generalizable methods, fine-tuning on clean data significantly boosts models' performance (See the first column in \tabref{tab:llffc} and \tabref{tab:blenderc}). However, this does not hold for corrupted data. Although MVSNeRF~\cite{chen2021mvsnerf} improves on all corruptions because of its generalization performance, IBRNet~\cite{wang2021ibrnet} drops on both datasets across several corruptions, with the largest degradation occurring at severity levels $> 1$. The main reason is that highly corrupted scenes disrupt the multi-view consistency that NeRF training relies on, and fine-tuning on such data causes the pretrained network to overfit incorrect geometries. An example can be found in \figref{fig:rendering}.

\section{Discussion}
In this section, we discuss various design and training details in NeRF-based systems and explore how they affect model robustness.

\topic{Feature encoding.} As mentioned in Section \ref{subsec:genres}, generalizable methods include an encoder $\bm{F}$ for image feature extraction. However, different existing methods have different encoder designs. For example, IBRNet~\cite{wang2021ibrnet} uses a U-Net structure containing several downsampling residual blocks and upsampling layers, resulting in 7.96 GFlops and 8.92M parameters, while MVSNeRF~\cite{chen2021mvsnerf} only has a few convolutional downsampling layers with 484.5 MFlops and 42.26k parameters. It is important to understand whether the choice of the encoder impacts both the quality of novel view synthesis and the robustness of generalizable methods. The number of parameters also has a direct impact on total training time, as volume rendering is already quite time-intensive.

Therefore, we present additional control studies on IBRNet~\cite{wang2021ibrnet} and MVSNeRF~\cite{chen2021mvsnerf} with encoders of different design choices by decreasing the channels of IBRNet's encoder or residual blocks (Please found the architecture details in the supplementary). We re-trained both methods on the IBRNet Collected and LLFF released scenes. Since the direct inference performance of MVSNeRF~\cite{chen2021mvsnerf} remains poor, we continue fine-tuning each target testing scene. The results are reported on \tabref{tab:encoder}.

\begin{table}[htbp]
\centering
\resizebox{\linewidth}{!}{
\begin{tabular}{c|c|ccc|ccc}
\toprule[0.5pt]
& & \multicolumn{3}{c|}{IBRNet} & \multicolumn{3}{c}{MVSNeRF (ft)} \\ \midrule[0.5pt]
Encoder & \multicolumn{1}{c|}{Flops} & PSNR & SSIM & LPIPS & PSNR & SSIM & LPIPS \\ \midrule[0.5pt]
ResUNet & 7.96G & 20.82 & 0.593 & 0.421 & 20.55 & 0.702 & 0.453 \\ 
ResNet  & 2.62G & 20.97 & 0.593 & 0.422 & 20.53 & 0.701 & 0.454 \\
ResUNet-Small & 2.22G & 20.81 & 0.593 & 0.422 & 20.50 & 0.700 & 0.456 \\
ResUNet-Tiny & 1.09G & 20.84 & 0.594 & 0.420 & 20.52 & 0.700 & 0.454 \\
MVSNeRF & 0.48G & 20.76 & 0.586 & 0.424 & 20.58 & 0.702 & 0.453 \\
\bottomrule[0.5pt]
\end{tabular}
}
\caption{Robustness results with different encoder designs.}
\label{tab:encoder}
\end{table}

According to the results, surprisingly, the choices of encoder contribute marginally both to clean data and corruption data. However, training time does vary, \eg in IBRNet, the total time for training ResUNet-Tiny compared with ResUNet-Big is reduced by more than 60\%. This suggests that only low-level features are needed for generalizable models in current frameworks, and the results might be more related to the feature aggregation part.

\topic{Patch-based sampling.} Recall that NeRF-based methods randomly sample a batch of rays in each training iteration and compare their predicted color with ground truth, which ensures ray diversity while training. Here, we explore another patch-based sampling strategy in which we sample $m$ numbers of $n \times n$ image patches instead, and $m \times n \times n$ equals the number of rays per iteration. We experiment with $n = 2, 4$ on NeRF~\cite{mildenhall2022nerf} and MVSNeRF~\cite{chen2021mvsnerf}, and find that it drops performance on clean data, but improves the absolute robustness on Fog ($0.1$ and $0.2$ PSNR increase respectively). With $n=2$, they are also more relatively robust.

\topic{Data augmentation.} We study if data-augmentation techniques help to improve NeRF-based methods' resistance to visual corruption. We sought image data augmentation at generalizable methods originally designed for classification to help the encoders extract more robust features. However, most augmentations disturb the pixel values and even corrupt the semantic information, making it impossible to train with NeRF-based methods. We test on Augmix~\cite{hendrycks2019augmix} that offer minimal deviation on the original image. We augment each image of a training scene with the same set of hyperparameters and train IBRNet~\cite{wang2021ibrnet}, MVSNeRF~\cite{chen2021mvsnerf} and GPNR~\cite{suhail2022generalizable}, and fail to observe improvements upon these methods, for example, IBRNet trained with augmentation have a 0.35 and 0.13 drop in PSNR for clean and corrupted inputs. The main reason those image-based data augmentation offers limited help is that the augmented scenes fail to correspond to a real physical 3D scene, so NeRF-based methods cannot find enough cross-view information to be trained with. We also removed the random crop and random flip operation originally in IBRNet, and found a performance drop in both clean and corrupted data (PSNR decreases by 0.17 and 0.15, respectively).

\topic{Pose estimation.} In the real-world scenario, structure-from-motion algorithms would be applied to estimate camera poses for each input image. It often fails with sparse or textureless inputs, let alone corrupted images that are difficult to find enough correspondence. Therefore, as part of our study, we also study how image corruption would influence bundle adjustment. We tested the commonly used open-source COLMAP framework~\cite{schonberger2016structure} on the real-world dataset LLFF-C. During the experiment, we found that COLMAP output is not fully deterministic and sometimes would fail even on dense clean images. Therefore, we try for each set of images a maximum of 5 times and stop when successful. If it fails all of them, the estimation is deemed unsuccessful.

For pose estimation, we are interested in whether a set of images for a 3D scene can be successfully calibrated. Following \cite{schonberger2016structure}, we use \textit{the number of triangulated points, mean track length, the number of images registered}, and \textit{the number of observations per image} as the metrics. Part of the results are shown in \tabref{tab:sfm}, and the rest can be found in the supplementary materials.

\begin{table}[htbp]
\centering
\resizebox{\linewidth}{!}{
\begin{tabular}{c|ccc|ccc}
\toprule[0.5pt]
& \multicolumn{3}{c|}{Num. Points Registered} & \multicolumn{3}{c}{Mean Track Length} \\ \midrule[0.5pt]
\diagbox{Ratio}{Severity} & 1 & 2 & 3 & 1 & 2 & 3\\ \midrule[0.5pt]
Clean Data (0\%) & \multicolumn{3}{c|}{4086.00} & \multicolumn{3}{c}{10.06} \\ \midrule[0.5pt]
10\% & 3948.53 & 3817.69 & 3821.90 & 7.90 & 7.80 & 7.81 \\
20\% & 3798.23 & 3796.88 & 3422.49 & 7.80 & 7.80 & 7.31 \\
50\% & 3663.30 & 3410.06 & 2828.70 & 7.21 & 6.95 & 6.18 \\
80\% & 3281.86 & 2661.33 & 2092.42 & 6.70 & 5.96 & 5.21 \\
100\% & 3087.33 & 2227.38 & 1434.42 & 6.90 & 5.81 & 4.59 \\
 \bottomrule[0.5pt]
\end{tabular}
}
\caption{Reconstruction results on the LLFF-C dataset at different corruption severity and ratio.}
\label{tab:sfm}
\end{table}

From \tabref{tab:sfm}, we can see that both severity and ratio negatively affect the reconstruction quality. When a small ratio of images is corrupted (10\%), the reconstruction is nearly not disturbed, and increases in the severity do not affect the results. However, with a higher corruption ratio, both metrics degraded quickly, and severity began to play a role. This suggests that when only a limited number of inputs are corrupted, camera poses can still be accurately estimated. However, the reconstruction system also becomes unreliable with severed corruptions, warranting pose correction modules in NeRF-based systems.

\section{Conclusion}
\label{sec:conclusion}
In this paper, we present the first benchmark for evaluating the robustness of NeRF-based methods, which was made possible by introducing two new datasets containing corrupted 3D scenes of several severity levels. To succeed in this benchmark, a system must have the ability to recover the physical world despite encountering different types of unseen corruption. Our results show that existing methods suffer significant performance drops in a manner different than recognition models, and standard image-based augmentation offers limited improvements. For generalizable methods, the feature encoder for current architectures contributes little to the robustness.
Our findings offer valuable insights into the robustness of NeRF-based models. We hope this will inspire future research towards developing more robust NeRF systems for real-world applications.

\section*{Acknowledgement}
This work is supported by a gift from Open Philanthropy, TPU Research Cloud (TRC) program, and Google Cloud Research Credits program.

%%%%%%%%% REFERENCES
{\small
\bibliographystyle{ieee_fullname}
\bibliography{egbib}
}

\appendix
\section{More Results}
\subsection{CM and RmCM}
$\mathrm{CM_{SSIM}}$ for LLFF-C and Blender-C can be found in \tabref{tab:llffc} and \tabref{tab:blenderc}. $\mathrm{RmCM}$ for PSNR, SSIM and LPIPS can be found from \tabref{tab:llffcrmcmpsnr} to \tabref{tab:blendercrmcmlpips}. We can see that most methods have a more than 25\% performance drop in PSNR both at LLFF-C and Blender-C.

\subsection{Pose Estimation}
More results for pose estimation are shown in \tabref{tab:sfm}. We can see that both the number of images registered and the number of observations per image decrease with the ratio of corrupted inputs.

\subsection{Patch Sampling}
We experimented with patch sampling on NeRF and MVSNeRF (See \tabref{tab:patch}). Generally, the absolute performance drops with larger patches, but the relative robustness increases.

\subsection{Finetuning on different encoders}
We further fine-tuned IBRNet with two encoders: ResUNet and ResUNet-Tiny. Same to direct inference, results showed that ResUNet ($27.64/0.891/0.114$ for PSNR/SSIM/LPIPS) is slightly better when finetuned on clean data ($27.52/0.888/0.117$ for PSNR/SSIM/LPIPS), but the robustness of ResUNet ($21.82/0.656/0.380$ for PSNR/SSIM/LPIPS) on corrupted data doesn't differ much with ResUNet-Tiny ($21.79/0.653/0.384$ for PSNR/SSIM/LPIPS).

\begin{table}
\centering
\resizebox{\linewidth}{!}{
\begin{tabular}{c|ccc|ccc}
\toprule[0.5pt]
& \multicolumn{3}{c|}{Num. Imaged Registered} & \multicolumn{3}{c}{Num. Observations Per Image} \\ \midrule[0.5pt]
\diagbox{Ratio}{Severity} & 1 & 2 & 3 & 1 & 2 & 3\\ \midrule[0.5pt]
Clean Data (0\%) & \multicolumn{3}{c|}{38.125} & \multicolumn{3}{c}{1006.75} \\ \midrule[0.5pt]
10\% & 34.982 & 34.777 & 34.830 & 976.73 & 924.87 & 947.57 \\
20\% & 34.741 & 35.652 & 33.946 & 912.45 & 908.74 & 789.07 \\
50\% & 33.938 & 35.813 & 33.107 & 826.90 & 699.68 & 568.39 \\
80\% & 32.232 & 33.375 & 31.741 & 720.71 & 497.82 & 366.10 \\
100\% & 32.241 & 31.821 & 27.634 & 679.31 & 420.37 & 261.68 \\
 \bottomrule[0.5pt]
\end{tabular}
}
\caption{Reconstruction results on the LLFF-C dataset at different corruption severity and ratio.}
\label{tab:sfm}
\end{table}

\section{Dataset}
\begin{figure*}[htbp]
    \centering
    \includegraphics[width=1.0\linewidth]{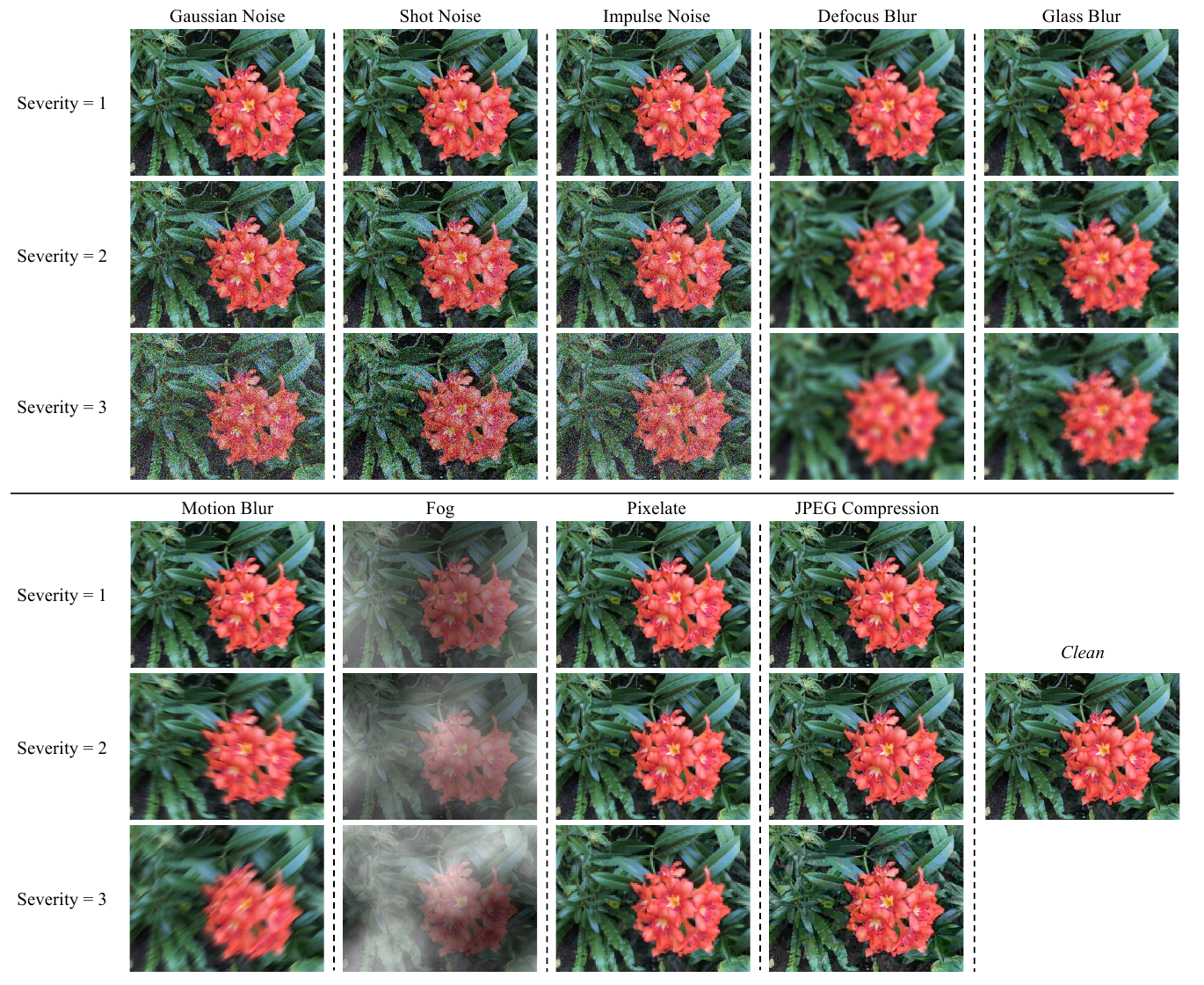}
    \caption{More image samples on LLFF-C}
    \label{fig:suppllffc}
\end{figure*}

\begin{figure*}[htbp]
    \centering
    \includegraphics[width=1.0\linewidth]{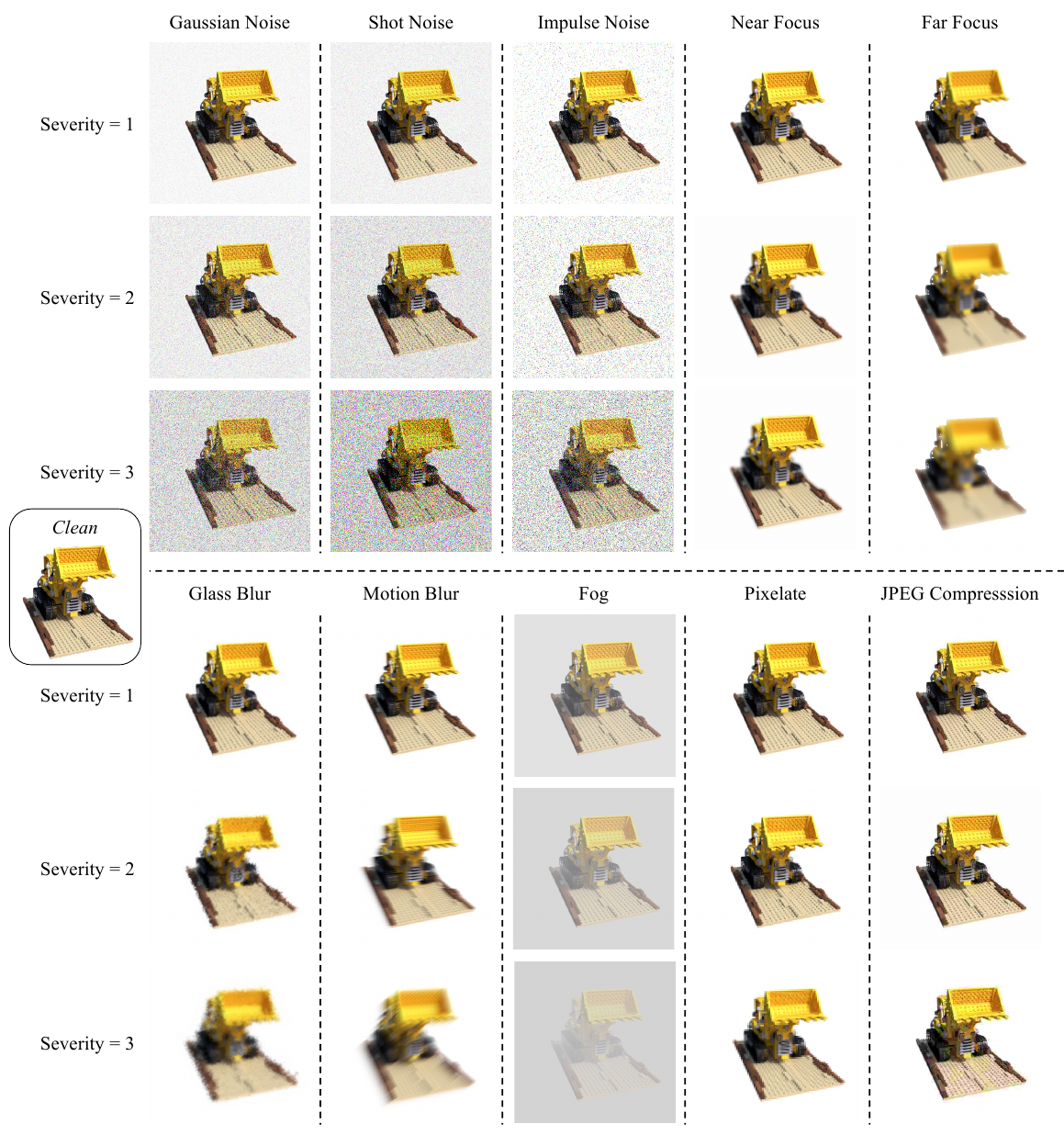}
    \caption{More image samples on LLFF-C}
    \label{fig:suppblenderc}
\end{figure*}

We construct LLFF-C at the resolution of $504 \times 378$, and Blender-C at the resolution of $400 \times 400$, mainly to save training time. See \figref{fig:suppllffc} and \figref{fig:suppblenderc} for a full set of corruption examples.

\section{Details for Each Method}
\topic{NeRF~\cite{mildenhall2020nerf}, MipNeRF~\cite{barron2021mip}} We trained scenes in both datasets for 200,000 iterations.

\topic{IBRNet~\cite{wang2021ibrnet}} We fine-tuned each scene in the LLFF-C and Blender-C datasets for 45,000 steps and 15,000 steps respectively.

\topic{MVSNeRF~\cite{chen2021mvsnerf}} requires the height and width of the image resolution divisible by 32, so the inputs images in LLFF-C and Blender-C are resized to $480 \times 320$ and $384 \times 384$ respectively. The metric evaluation is done at the resized resolution. MVSNeRF uses 15 images for training and another 4 images for testing in the LLFF dataset, but we followed the original train-test split for LLFF as in the original NeRF. As for the Blender-C dataset, we use the target training set and target testing set partition in MVSNeRF since we found MVSNeRF performs badly on the standard one. We fine-tuned each scene for 1 epoch.

\topic{Plenoxel~\cite{fridovich2022plenoxels}} We use the official code directly for optimization and evaluation.

\topic{GPNR~\cite{suhail2022generalizable}} Since there is no official checkpoint available, we trained GPNR on the IBRNet collected dataset for 150,000 steps and tested it on LLFF-C.

\section{Encoder Architecture}
\begin{figure}[htbp]
    \centering
    \includegraphics[width=1.0\linewidth]{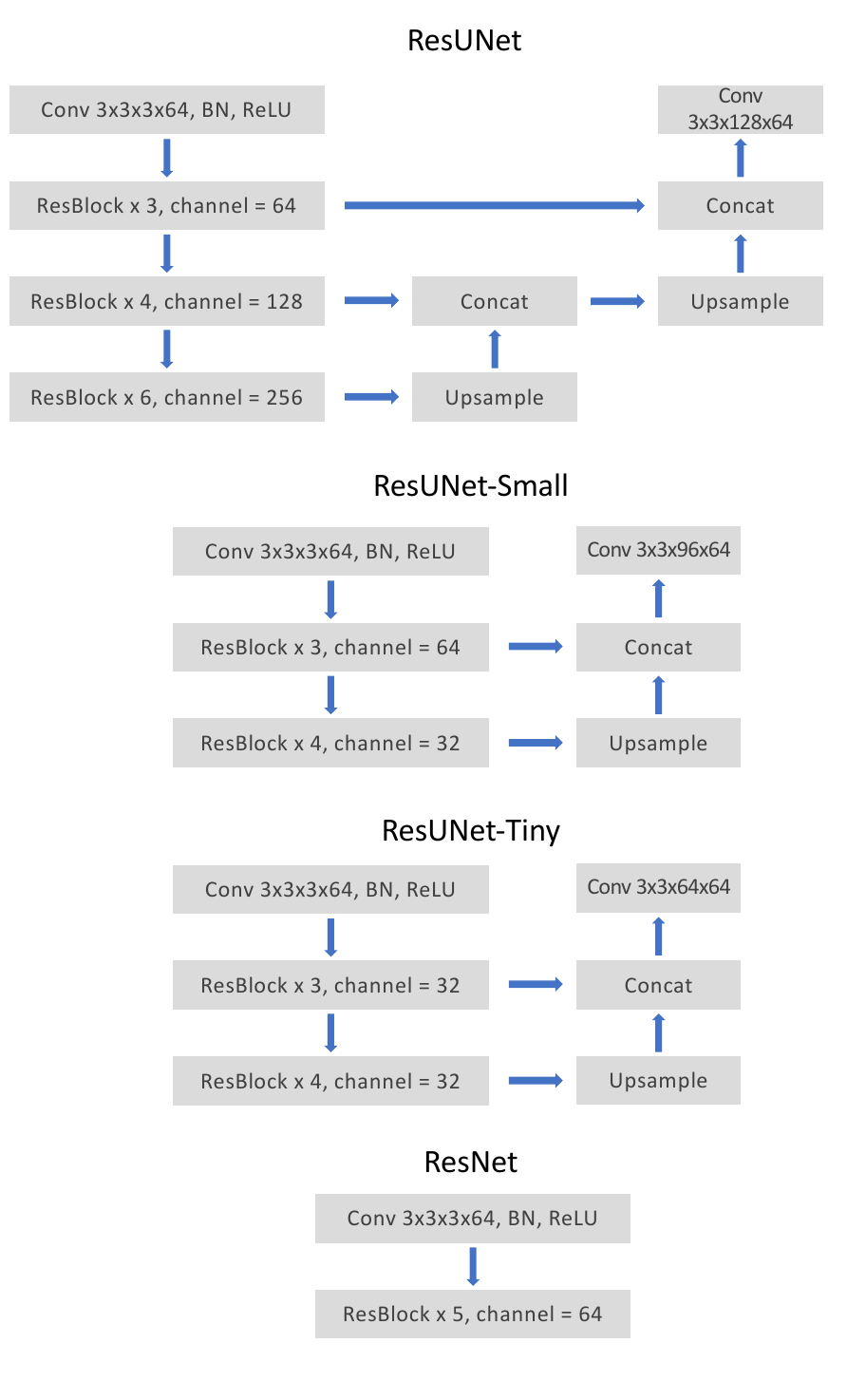}
    \caption{More image samples on LLFF-C}
    \label{fig:encoder}
\end{figure}

The feature encoders we experimented with for generalizable methods can be found in \figref{fig:encoder}.

\begin{table*}[htbp]
\resizebox{\linewidth}{!}{
\begin{tabular}{l|l|ccc|ccc|c|cc}
\toprule[0.5pt]
 & & \multicolumn{3}{c|}{Noise} & \multicolumn{3}{c|}{Blur} & \multicolumn{1}{c|}{Weather} & \multicolumn{2}{c}{Digital} \\
\multicolumn{1}{c|}{}  & \multicolumn{1}{c|}{Clean} & Gauss. & Shot & Impulse & Defocus & Glass & Motion & Fog & Pixel & JPEG \\ \midrule[0.5pt]
NeRF (patch=2) & 27.40/0.164 & 24.16/0.308 & 23.58/0.311 & 23.90/0.307 & 20.83/0.507 & 22.24/0.378 & 19.35/0.483 & 11.99/0.545 & 24.23/0.306 & 25.20/0.293 \\
NeRF (patch=4) & 26.92/0.184 & 23.98/0.313 & 23.45/0.317 & 23.78/0.312 & 20.84/0.512 & 22.19/0.385 & 19.47/0.489 & 11.89/0.635 & 23.31/0.299 & 24.99/0.301 \\
MVSNeRF (patch=2) & 23.24/0.267 & 21.13/0.457 & 20.84/0.458 & 21.06/0.457 & 20.36/0.523 & 21.29/0.425 & 19.24/0.504 & 13.35/0.610 & 22.23/0.373 & 22.61/0.368 \\
MVSNeRF (patch=4) & 22.74/0.285 & 20.73/0.470 & 20.51/0.469 & 20.70/0.470 & 20.07/0.536 & 20.89/0.442 & 19.00/0.516 & 13.53/0.614 & 21.78/0.389 & 22.09/0.385
\\ \bottomrule[0.5pt]
\end{tabular}
}
\caption{PSNR/LPIPS results for clean and corrupted data on LLFF-C, ft indicates results after fine-tuning for generalizable methods.}
\label{tab:patch}
\end{table*}

\begin{table*}[htbp]
\resizebox{\linewidth}{!}{
\begin{tabular}{l|l|ccc|ccc|c|cc}
\toprule[0.5pt]
 & & \multicolumn{3}{c|}{Noise} & \multicolumn{3}{c|}{Blur} & \multicolumn{1}{c|}{Weather} & \multicolumn{2}{c}{Digital} \\
\multicolumn{1}{c|}{}  & \multicolumn{1}{c|}{Clean} & Gauss. & Shot & Impulse & Defocus & Glass & Motion & Fog & Pixel & JPEG \\ \midrule[0.5pt]
NeRF & 0.876 & 0.769 & 0.773 & 0.768 & 0.521 & 0.661 & 0.498 & 0.398 & 0.750 & 0.797 \\
MipNeRF & 0.874 & 0.722 & 0.719 & 0.718 & 0.521 & 0.658 & 0.487 & 0.396 & 0.746 & 0.698 \\
Plenoxel & 0.900 & 0.481 & 0.481 & 0.458 & 0.529 & 0.659 & 0.527 & 0.249 & 0.747 & 0.788 \\
MVSNeRF & 0.629 & 0.546 & 0.543 & 0.542 & 0.549 & 0.588 & 0.523 & 0.487 & 0.616 & 0.613 \\
MVSNeRF (ft) & 0.842 & 0.735 & 0.743 & 0.736 & 0.645 & 0.729 & 0.635 & 0.510 & 0.777 & 0.806 \\
IBRNet & 0.844 & 0.572 & 0.575 & 0.598 & 0.508 & 0.637 & 0.516 & 0.473 & 0.722 & 0.748 \\
IBRNet (ft) & 0.893 & 0.745 & 0.577 & 0.740 & 0.491 & 0.631 & 0.517 & 0.429 & 0.685 & 0.766 \\
GPNR & 0.813 & 0.576 & 0.575 & 0.581 & 0.500 & 0.616 & 0.497 & 0.376 & 0.699 & 0.731
\\ \bottomrule[0.5pt]
\end{tabular}
}
\caption{SSIM results for clean and corrupted data on LLFF-C, ft indicates results after fine-tuning for generalizable methods.}
\label{tab:llffc}
\end{table*}

\begin{table*}[htbp]
\resizebox{\linewidth}{!}{
\begin{tabular}{l|l|ccc|cccc|c|cc}
\toprule[0.5pt]
 & & \multicolumn{3}{c|}{Noise} & \multicolumn{4}{c|}{Blur} & \multicolumn{1}{c|}{Weather} & \multicolumn{2}{c}{Digital} \\
\multicolumn{1}{c|}{}  & \multicolumn{1}{c|}{Clean} & Gauss. & Shot & Impulse & Near Focus & Far Focus & Glass & Motion & Fog & Pixel & JPEG \\ \midrule[0.5pt]
NeRF         & 0.952 & 0.919 & 0.912 & 0.918 & 0.912 & 0.881 & 0.832 & 0.812 & 0.836 & 0.906 & 0.925 \\
MipNeRF      & 0.857 & 0.849 & 0.717 & 0.759 & 0.808 & 0.848 & 0.841 & 0.694 & 0.796 & 0.850 & 0.823 \\
Plenoxel     & 0.966 & 0.445 & 0.338 & 0.538 & 0.928 & 0.896 & 0.828 & 0.811 & 0.750 & 0.904 & 0.923 \\
MVSNeRF      & 0.857 & 0.849 & 0.717 & 0.759 & 0.808 & 0.848 & 0.841 & 0.694 & 0.796 & 0.850 & 0.823 \\
MVSNeRF (ft) & 0.883 & 0.806 & 0.781 & 0.874 & 0.818 & 0.876 & 0.877 & 0.782 & 0.805 & 0.879 & 0.850 \\
IBRNet       & 0.926 & 0.667 & 0.587 & 0.616 & 0.896 & 0.867 & 0.820 & 0.815 & 0.824 & 0.889 & 0.898 \\
IBRNet (ft)  & 0.951 & 0.685 & 0.635 & 0.631 & 0.893 & 0.857 & 0.835 & 0.817 & 0.816 & 0.878 & 0.875 \\
\bottomrule[0.5pt]
\end{tabular}
}
\caption{SSIM results for clean and corrupted data on Blender-C, ft indicates results after fine-tuning for generalizable methods.}
\label{tab:blenderc}
\end{table*}

\begin{table*}[htbp]
\resizebox{\linewidth}{!}{
\begin{tabular}{l|l|ccc|ccc|c|cc}
\toprule[0.5pt]
 & & \multicolumn{3}{c|}{Noise} & \multicolumn{3}{c|}{Blur} & \multicolumn{1}{c|}{Weather} & \multicolumn{2}{c}{Digital} \\
\multicolumn{1}{c|}{}  & \multicolumn{1}{c|}{Avg.} & Gauss. & Shot & Impulse & Defocus & Glass & Motion & Fog & Pixel & JPEG \\ \midrule[0.5pt]
NeRF         & 0.21 & 0.12 & 0.14 & 0.13 & 0.25 & 0.31 & 0.20 & 0.57 & 0.12 & 0.09 \\
MipNeRF      & 0.23 & 0.13 & 0.16 & 0.14 & 0.25 & 0.31 & 0.20 & 0.57 & 0.13 & 0.20 \\
Plenoxel     & 0.27 & 0.25 & 0.26 & 0.35 & 0.26 & 0.30 & 0.20 & 0.55 & 0.13 & 0.10 \\
MVSNeRF      & 0.05 & 0.01 & 0.03 & 0.02 & 0.01 & 0.04 & 0.02 & 0.24 & 0.04 & 0.02 \\
MVSNeRF (ft) & 0.14 & 0.10 & 0.12 & 0.11 & 0.14 & 0.19 & 0.10 & 0.45 & 0.05 & 0.03 \\
IBRNet       & 0.19 & 0.15 & 0.17 & 0.14 & 0.20 & 0.25 & 0.15 & 0.47 & 0.09 & 0.07 \\
IBRNet (ft)  & 0.24 & 0.14 & 0.23 & 0.14 & 0.29 & 0.30 & 0.21 & 0.52 & 0.18 & 0.11 \\
GPNR         & 0.17 & 0.12 & 0.14 & 0.13 & 0.17 & 0.22 & 0.13 & 0.48 & 0.07 & 0.05
\\ \bottomrule[0.5pt]
\end{tabular}
}
\caption{RmCM results for PSNR on LLFF-C, ft indicates results after fine-tuning for generalizable methods.}
\label{tab:llffcrmcmpsnr}
\end{table*}

\begin{table*}[htbp]
\resizebox{\linewidth}{!}{
\begin{tabular}{l|l|ccc|ccc|c|cc}
\toprule[0.5pt]
 & & \multicolumn{3}{c|}{Noise} & \multicolumn{3}{c|}{Blur} & \multicolumn{1}{c|}{Weather} & \multicolumn{2}{c}{Digital} \\
\multicolumn{1}{c|}{}  & \multicolumn{1}{c|}{Avg.} & Gauss. & Shot & Impulse & Defocus & Glass & Motion & Fog & Pixel & JPEG \\ \midrule[0.5pt]
NeRF         & 0.25 & 0.12 & 0.12 & 0.12 & 0.41 & 0.25 & 0.43 & 0.55 & 0.14 & 0.09 \\
MipNeRF      & 0.28 & 0.17 & 0.18 & 0.18 & 0.40 & 0.25 & 0.44 & 0.55 & 0.15 & 0.20 \\
Plenoxel     & 0.39 & 0.47 & 0.47 & 0.49 & 0.41 & 0.27 & 0.41 & 0.72 & 0.17 & 0.12 \\
MVSNeRF      & 0.12 & 0.13 & 0.14 & 0.14 & 0.13 & 0.07 & 0.17 & 0.23 & 0.02 & 0.03 \\
MVSNeRF (ft) & 0.17 & 0.13 & 0.12 & 0.13 & 0.23 & 0.13 & 0.25 & 0.39 & 0.08 & 0.04 \\
IBRNet       & 0.30 & 0.32 & 0.32 & 0.29 & 0.40 & 0.25 & 0.39 & 0.44 & 0.15 & 0.11 \\
IBRNet (ft)  & 0.31 & 0.17 & 0.35 & 0.17 & 0.45 & 0.29 & 0.42 & 0.52 & 0.23 & 0.14 \\
GPNR         & 0.30 & 0.29 & 0.29 & 0.29 & 0.39 & 0.24 & 0.39 & 0.54 & 0.14 & 0.10
\\ \bottomrule[0.5pt]
\end{tabular}
}
\caption{RmCM results for SSIM on LLFF-C, ft indicates results after fine-tuning for generalizable methods.}
\label{tab:llffcrmcmssim}
\end{table*}

\begin{table*}[htbp]
\resizebox{\linewidth}{!}{
\begin{tabular}{l|l|ccc|ccc|c|cc}
\toprule[0.5pt]
 & & \multicolumn{3}{c|}{Noise} & \multicolumn{3}{c|}{Blur} & \multicolumn{1}{c|}{Weather} & \multicolumn{2}{c}{Digital} \\
\multicolumn{1}{c|}{}  & \multicolumn{1}{c|}{Avg.} & Gauss. & Shot & Impulse & Defocus & Glass & Motion & Fog & Pixel & JPEG \\ \midrule[0.5pt]
NeRF         & 1.48 & 0.97 & 0.99 & 0.97 & 2.36 & 1.46 & 2.20 & 2.50 & 1.00 & 0.91 \\
MipNeRF      & 1.47 & 1.13 & 1.17 & 1.14 & 2.20 & 1.38 & 2.18 & 2.42 & 0.98 & 0.61 \\
Plenoxel     & 3.50 & 3.80 & 3.86 & 3.94 & 4.08 & 2.72 & 3.63 & 5.59 & 1.88 & 2.00 \\
MVSNeRF      & 0.29 & 0.33 & 0.35 & 0.35 & 0.41 & 0.22 & 0.36 & 0.39 & 0.10 & 0.14 \\
MVSNeRF (ft) & 0.84 & 0.81 & 0.81 & 0.81 & 1.11 & 0.67 & 1.01 & 1.48 & 0.45 & 0.43 \\
IBRNet       & 1.67 & 1.86 & 1.88 & 1.75 & 2.25 & 1.48 & 1.87 & 1.87 & 0.97 & 1.14 \\
IBRNet (ft)  & 2.57 & 1.85 & 3.01 & 1.87 & 3.59 & 2.51 & 3.05 & 3.46 & 2.03 & 1.77 \\
GPNR         & 1.18 & 1.28 & 1.29 & 1.26 & 1.52 & 0.98 & 1.29 & 1.70 & 0.62 & 0.64
\\ \bottomrule[0.5pt]
\end{tabular}
}
\caption{RmCM results for LPIPS on LLFF-C, ft indicates results after fine-tuning for generalizable methods.}
\label{tab:llffcrmcmlpips}
\end{table*}

\begin{table*}[htbp]
\resizebox{\linewidth}{!}{
\begin{tabular}{l|l|ccc|cccc|c|cc}
\toprule[0.5pt]
 & & \multicolumn{3}{c|}{Noise} & \multicolumn{4}{c|}{Blur} & \multicolumn{1}{c|}{Weather} & \multicolumn{2}{c}{Digital} \\
\multicolumn{1}{c|}{}  & \multicolumn{1}{c|}{Avg.} & Gauss. & Shot & Impulse & Near Focus & Far Focus & Glass & Motion & Fog & Pixel & JPEG \\ \midrule[0.5pt]
NeRF         & 0.26 & 0.29 & 0.38 & 0.23 & 0.13 & 0.20 & 0.27 & 0.33 & 0.50 & 0.13 & 0.09 \\
MipNeRF      & 0.29 & 0.32 & 0.41 & 0.25 & 0.16 & 0.22 & 0.33 & 0.37 & 0.53 & 0.17 & 0.11 \\
Plenoxel     & 0.31 & 0.38 & 0.46 & 0.32 & 0.16 & 0.22 & 0.32 & 0.36 & 0.53 & 0.19 & 0.13 \\
MVSNeRF      & 0.12 & 0.02 & 0.27 & 0.20 & 0.12 & 0.01 & 0.03 & 0.21 & 0.28 & 0.02 & 0.07 \\
MVSNeRF (ft) & 0.15 & 0.24 & 0.31 & 0.03 & 0.19 & 0.02 & 0.02 & 0.22 & 0.38 & 0.01 & 0.08 \\
IBRNet       & 0.20 & 0.25 & 0.33 & 0.21 & 0.08 & 0.12 & 0.20 & 0.24 & 0.44 & 0.08 & 0.06 \\
IBRNet (ft)  & 0.28 & 0.30 & 0.33 & 0.25 & 0.17 & 0.24 & 0.28 & 0.33 & 0.49 & 0.20 & 0.20 \\
\bottomrule[0.5pt]
\end{tabular}
}
\caption{RmCM results for PSNR on Blender-C, ft indicates results after fine-tuning for generalizable methods.}
\label{tab:blendercrmcmpsnr}
\end{table*}

\begin{table*}[htbp]
\resizebox{\linewidth}{!}{
\begin{tabular}{l|l|ccc|cccc|c|cc}
\toprule[0.5pt]
 & & \multicolumn{3}{c|}{Noise} & \multicolumn{4}{c|}{Blur} & \multicolumn{1}{c|}{Weather} & \multicolumn{2}{c}{Digital} \\
\multicolumn{1}{c|}{}  & \multicolumn{1}{c|}{Avg.} & Gauss. & Shot & Impulse & Near Focus & Far Focus & Glass & Motion & Fog & Pixel & JPEG \\ \midrule[0.5pt]
NeRF         & 0.07 & 0.03 & 0.04 & 0.04 & 0.04 & 0.07 & 0.13 & 0.15 & 0.12 & 0.05 & 0.03 \\
MipNeRF      & 0.07 & 0.01 & 0.16 & 0.11 & 0.06 & 0.01 & 0.02 & 0.19 & 0.07 & 0.01 & 0.04 \\
Plenoxel     & 0.24 & 0.54 & 0.65 & 0.44 & 0.04 & 0.07 & 0.14 & 0.16 & 0.22 & 0.06 & 0.04 \\
MVSNeRF      & 0.07 & 0.01 & 0.16 & 0.11 & 0.06 & 0.01 & 0.02 & 0.19 & 0.07 & 0.01 & 0.04 \\
MVSNeRF (ft) & 0.05 & 0.09 & 0.12 & 0.01 & 0.07 & 0.01 & 0.01 & 0.11 & 0.09 & 0.01 & 0.04 \\
IBRNet       & 0.15 & 0.28 & 0.37 & 0.34 & 0.03 & 0.06 & 0.11 & 0.12 & 0.11 & 0.04 & 0.03 \\
IBRNet (ft)  & 0.17 & 0.28 & 0.33 & 0.34 & 0.06 & 0.10 & 0.12 & 0.14 & 0.14 & 0.08 & 0.08 \\
\bottomrule[0.5pt]
\end{tabular}
}
\caption{RmCM results for SSIM on Blender-C, ft indicates results after fine-tuning for generalizable methods.}
\label{tab:blendercrmcmssim}
\end{table*}

\begin{table*}[htbp]
\resizebox{\linewidth}{!}{
\begin{tabular}{l|l|ccc|cccc|c|cc}
\toprule[0.5pt]
 & & \multicolumn{3}{c|}{Noise} & \multicolumn{4}{c|}{Blur} & \multicolumn{1}{c|}{Weather} & \multicolumn{2}{c}{Digital} \\
\multicolumn{1}{c|}{}  & \multicolumn{1}{c|}{Avg.} & Gauss. & Shot & Impulse & Near Focus & Far Focus & Glass & Motion & Fog & Pixel & JPEG \\ \midrule[0.5pt]
NeRF         & 1.41 & 1.29  & 1.91  & 1.34  & 0.78 & 1.21 & 1.94 & 2.26 & 1.81  & 0.93 & 0.63 \\
MipNeRF      & 1.58 & 1.44  & 2.45  & 1.91  & 0.62 & 0.99 & 2.40 & 2.47 & 2.01  & 0.92 & 0.60 \\
Plenoxel     & 7.51 & 14.82 & 16.16 & 12.11 & 1.85 & 2.29 & 4.96 & 5.40 & 12.56 & 2.70 & 2.26 \\
MVSNeRF      & 0.46 & 0.08  & 1.10  & 0.94  & 0.26 & 0.10 & 0.12 & 1.08 & 0.62  & 0.09 & 0.24 \\
MVSNeRF (ft) & 0.73 & 1.58  & 1.82  & 0.05  & 0.53 & 0.11 & 0.09 & 1.70 & 1.04  & 0.09 & 0.28 \\
IBRNet       & 1.13 & 2.43  & 2.75  & 2.49  & 0.01 & 0.48 & 0.89 & 0.78 & 0.66  & 0.39 & 0.39 \\
IBRNet (ft)  & 2.48 & 4.96  & 5.12  & 5.04  & 0.55 & 1.43 & 1.50 & 1.60 & 2.16  & 1.18 & 1.30 \\
\bottomrule[0.5pt]
\end{tabular}
}
\caption{RmCM results for LPIPS on Blender-C, ft indicates results after fine-tuning for generalizable methods.}
\label{tab:blendercrmcmlpips}
\end{table*}

\end{document}